%% file: root.tex
\documentclass[letterpaper, 10 pt, conference]{ieeeconf}  % Comment this line out if you need a4paper

\IEEEoverridecommandlockouts                              % This command is only needed if 
\usepackage{amsmath,amssymb,bm}
\usepackage{graphicx}
\usepackage{booktabs,multirow,array}
\usepackage{xcolor}
\usepackage{tikz}
\usetikzlibrary{positioning,arrows.meta,fit}
\usepackage[hidelinks]{hyperref}
\usepackage[capitalise,noabbrev]{cleveref}
\usepackage{cite}
\crefname{figure}{Fig.}{Figs.}\Crefname{figure}{Fig.}{Figs.}
\crefname{equation}{}{}\Crefname{equation}{Eq.}{Eqs.}
\creflabelformat{equation}{(#2#1#3)}
\crefname{section}{Sec.}{Secs.}\Crefname{section}{Sec.}{Secs.}
\usepackage{bm}   % bold math (if not already loaded)

\newcommand{\vect}[1]{\bm{#1}}

\title{\LARGE \bf
Old-School Planners Drive (Sometimes) Better Than Learned Ones
}

\author{Andreas Look$^{1}$ % <-this % stops a space
\thanks{$^{1}$Andreas Look is with the Faculty of Mechanical and Automotive Engineering,
        Coburg University of Applied Sciences and Arts, Coburg, Germany
        {\tt\small andreas.look@hs-coburg.de}}%
}

\begin{document}

\maketitle
\thispagestyle{empty}
\pagestyle{empty}

\input{sections/00_abstract}

\input{sections/01_introduction}
\input{sections/02_related}
\input{sections/03_method}
\input{sections/04_protocol}
\input{sections/05_results}
\input{sections/07_discussion}

\bibliographystyle{IEEEtran}
\bibliography{IEEEabrv,mybib}

\end{document}

%% file: sections/00_abstract.tex
\begin{abstract}
End-to-end driving models learn to plan from data, and hand-written planners are widely treated as a thing of the
past. We test whether this is justified: we implement sixteen textbook planners, each combining a path source (reference line, A*,
hybrid A*, RRT*) with a trajectory method (Frenet lattice, nonlinear optimization, $s$--$t$ search, dynamic window),
and evaluate them on three benchmarks: open loop on nuScenes, closed loop on nuPlan with ground-truth perception, and
photorealistic closed loop on HUGSIM. On HUGSIM, they replace only the planner of the end-to-end model UniAD and read
its frozen perception, without an HD map or other privileged information. On nuScenes and nuPlan, the best textbook
planners are competitive (L2 error of 0.735\,m against 0.947\,m for UniAD's planner; a nuPlan score of 79.3 against 76.1 for
the rule-based PDM-Closed on Test14-hard with reactive agents). On HUGSIM, nine of the sixteen score above UniAD's learned planner on the same
perception,. The best, a hybrid A* path followed by nonlinear trajectory optimization, raises
HUGSIM's driving score (HD-Score) from 32.0 to 41.8. The gain is
largest with standing and oncoming cars; against cars that steer at the ego, there is none. In 31 of the
182 runs in which UniAD's planner hits a car, the best textbook planner completes the route on the same perception:
these crashes are due to the planner, not the perception.
\end{abstract}

%% file: sections/01_introduction.tex
\section{Introduction}\label{sec:intro}

End-to-end driving models such as UniAD~\cite{hu2023uniad} and its
successors~\cite{ch01:jiang2023vad,ch01:weng2024paradrive,ch01:sun2024sparsedrive,ch01:liao2025diffusiondrive} map
raw sensor data to a trajectory and learn the planner from human driving logs. Research on these models mostly
compares them with other learned planners~\cite{ch01:liao2025diffusiondrive,li2024hydramdp,li2025wote}. Textbook
planners, i.e., rule-based methods such as hybrid A*~\cite{dolgov2010hybridastar}, Frenet lattices~\cite{werling2010lattice} and
optimization~\cite{ch01:ziegler2014bertha,fan2018apollo} are widely treated as a thing of the past and rarely appear
as baselines; NAVSIM~\cite{ch01:dauner2024navsim}, for instance, reports constant velocity and an ego-status MLP as its
simplest ones. Where rule-based planners do beat learned ones, as PDM-Closed~\cite{dauner2023pdm} did on
nuPlan~\cite{caesar2021nuplan}, they receive ground-truth objects and an HD map. The planner inside an end-to-end
model, in contrast, sees the world through learned, noisy perception. The closest test on such perception,
NeuroNCAP~\cite{ch01:ljungbergh2024neuroncap}, found that a simple braking rule on UniAD's detections already scores
above UniAD's planner, but tested no complete planner.

Two questions are therefore open. How well do textbook planners drive on today’s benchmarks? And do they still drive well when their input is the noisy perception of an end-to-end model instead of ground-truth objects and an HD map?
The second also concerns the model itself: in our runs, UniAD crashes in two thirds of the
scenarios of the photorealistic simulator HUGSIM~\cite{zhou2024hugsim}, and it is unknown whether its perception or its
planner is to blame.

\begin{figure}[t]
\centering
\includegraphics[width=\columnwidth]{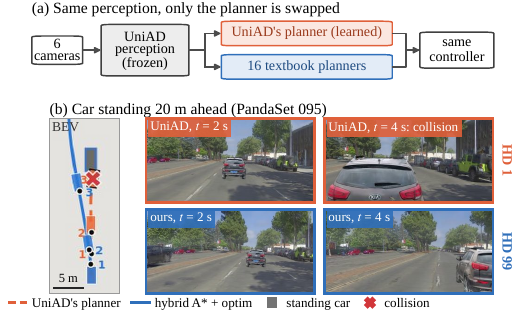}
\caption{(a)~On HUGSIM, UniAD's planner and sixteen textbook planners read the outputs of the same frozen UniAD
perception and return six waypoints to the same controller. (b)~A car stands still 20\,m ahead (numbered dots: ego positions over time). UniAD's learned
planner keeps driving and hits it after 4\,s (HD-Score 1 of 100). The best textbook planner, hybrid A* with trajectory
optimization (optim), treats the detected car as an obstacle and passes it on the left (HD-Score 99).}
\label{fig:teaser}
\end{figure}

We study sixteen textbook planners, each combining a path source with a trajectory method, on three benchmarks of
increasing realism: open loop on nuScenes~\cite{caesar2020nuscenes} with UniAD's perception, closed loop on nuPlan with ground-truth
perception, and closed loop on HUGSIM with the perception of UniAD. On HUGSIM, we keep the official UniAD checkpoint
frozen and replace only its planner (\cref{fig:teaser}). The textbook planners read UniAD's detected objects, forecasts and road masks, the
navigation command and the ego state, use no HD map or other privileged information, and return the same six
waypoints to the same controller, so any difference in driving is due to the planner. In summary, we contribute
sixteen textbook planners that run on the outputs of an end-to-end model, without an HD map,
and show:
\begin{itemize}
  \item On nuScenes and nuPlan, the best textbook planners are competitive: a Frenet lattice reaches
        an open-loop L2 error of 0.735\,m against 0.947\,m for UniAD's planner and comes within 3 points of
        PDM-Closed on nuPlan (\cref{sec:results}) on Val14 and outperforms it on Test14-hard.
  \item On UniAD's perception, nine of the sixteen drive better than UniAD's learned planner. The best raises the
        HD-Score from 32.0 to 41.8, above all but one of the learned models evaluated on the same scenarios
        (\cref{sec:hugsim}).
  \item The gain is largest with standing cars, which the path source drives around, and with oncoming cars, which
        the trajectory method checks against their forecasts. Against cars that steer at the ego, there is none.
\end{itemize}

%% file: sections/02_related.tex
\section{Related Work}\label{sec:related}

\textbf{End-to-end driving.} UniAD~\cite{hu2023uniad} chains modules for tracking, online mapping, motion forecasting
and occupancy prediction; a learned planner attends to their features, and a post-optimizer pushes the waypoints
away from predicted occupancy. Follow-up work changes the scene representation (VAD~\cite{ch01:jiang2023vad},
SparseDrive~\cite{ch01:sun2024sparsedrive}), runs the modules in parallel (PARA-Drive~\cite{ch01:weng2024paradrive}),
generates trajectories by diffusion (DiffusionDrive~\cite{ch01:liao2025diffusiondrive}), or scores many proposals with
heads distilled from human and rule-based teachers (Hydra-MDP~\cite{li2024hydramdp}). NMP, P3 and
MP3~\cite{ch01:zeng2019nmp,ch01:sadat2020p3,casas2021mp3} select sampled trajectories with learned costs, and
SafetyNet~\cite{vitelli2022safetynet} and InterFuser~\cite{shao2022interfuser} check or constrain a learned planner's
output with hand-written rules. In all of these, the component that decides the trajectory is learned or only checks
a learned one; we replace it with complete textbook planners.

\textbf{Why learned planners fail.} Behavior cloning needs synthesized perturbations to drive
robustly~\cite{ch01:bansal2019chauffeurnet} and suffers from dataset bias~\cite{ch01:codevilla2019limitations}.
Learned planners also pick up shortcuts: they copy the expert's previous action~\cite{wen2020copycat}, rely on the
ego status~\cite{ch01:cheng2024plantf,li2024egostatus} or depend on the target
point~\cite{ch01:jaeger2023hiddenbiases}.

\textbf{Rule-based planning and closed-loop benchmarks.} On nuPlan, PDM-Closed~\cite{dauner2023pdm} scores
IDM~\cite{treiber2000idm} proposals with a function that closely resembles the benchmark metric; the strongest recent
planners combine learning with rule-based selection or refinement~\cite{ch01:cheng2024pluto,diffusionplanner2025}.
All of these use ground-truth perception. NAVSIM~\cite{ch01:dauner2024navsim} evaluates end-to-end models in
non-reactive simulation, Bench2Drive~\cite{ch01:jia2024bench2drive} fully closed loop in CARLA, and
HUGSIM~\cite{zhou2024hugsim} in photorealistic closed loop on reconstructed real drives, where several end-to-end
models have been evaluated zero-shot~\cite{ch08:wajepa2026,drivezero2026}.

%% file: sections/03_method.tex
\section{Sixteen Textbook Planners}\label{sec:method}

\begin{figure*}[t]
\centering
\includegraphics[width=\textwidth]{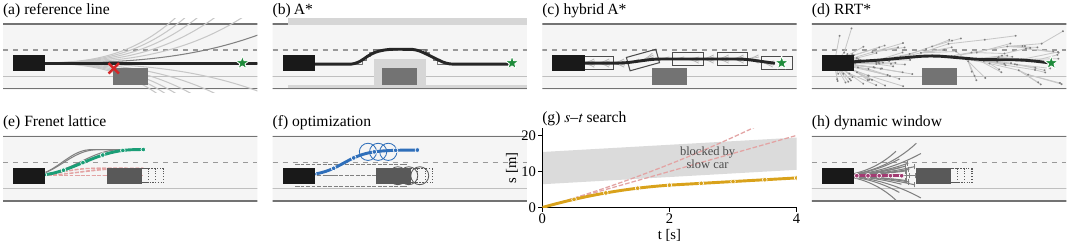}
\caption{Overview of path sources and trajectory methods on a synthetic street (black: ego; gray: other cars; green star: goal). Top, path
sources around a parked car: (a)~reference line, which follows the road and runs through the parked car (red cross); avoiding it is left to the trajectory method; (b)~A* on a grid with
blocked cells shaded; (c)~hybrid A* with bounded-curvature arcs; (d)~RRT*. Bottom, trajectory methods behind a slow
car (dots: future ego positions; dashed red: colliding candidates): (e)~lattice; (f)~optimization from several initial
guesses; (g)~$s$--$t$ search; (h)~dynamic window.}
\label{fig:methods}
\end{figure*}

Every planner follows the textbook split of motion planning~\cite{ch01:paden2016survey}: a
\emph{path source} decides where to drive, and a \emph{trajectory method} decides how fast to drive along that path and how far to leave
it. We implement four of each and combine them, which gives sixteen planners (\cref{fig:methods}).

\subsection{What the planners read from UniAD}\label{sec:interface}

In every cycle, a planner receives UniAD's tracked agents with motion forecasts over 6\,s, the predicted vehicle
occupancy, bird's-eye-view masks of the drivable area, the ego speed, acceleration and yaw
rate, and the navigation command (left, right, straight). It returns six waypoints at 0.5\,s spacing, the format of
UniAD's own plan.

Map-based planners take the route, goal, free space and speed limit from an HD map; we take them from UniAD's
outputs. A \emph{reference line} replaces the route: we fan out candidate curves from the ego and take the one that
stays longest on the predicted drivable mask among those that turn as the command says. The goal of all path sources is the point on the reference line just beyond what the ego can
reach in 3\,s, at most 50\,m ahead. The predicted
drivable mask replaces the free space, and, without a speed limit, the target speed is derived
from the current speed, with a tuned minimum.

Static agents become obstacles on the path source's grid, so the path goes around them. Moving agents are left to
the trajectory method, and so is a stopped vehicle that UniAD detects centered in the ego's lane and aligned with it,
which may only be waiting in traffic. All trajectory methods compare the planned ego position at each time step
with the forecast position of each agent at that time.

\subsection{Path sources: where to drive}\label{sec:paths}

The four path sources differ in how they get to the goal (\cref{fig:methods}a--d). The simplest uses the
\textbf{reference line} itself as the path and ignores obstacles.

\textbf{A*}~\cite{hart1968astar} searches a 0.5\,m grid for the cheapest path of the ego's center. Cells off the
drivable mask or within half the ego width of a static obstacle are blocked.

\textbf{Hybrid A*}~\cite{dolgov2010hybridastar} extends A* to paths the car can drive: it searches over positions and
headings with short arcs of bounded curvature and checks the full car footprint against obstacles.

\textbf{RRT*}~\cite{karaman2011rrtstar} grows and rewires a tree of random samples, biased towards the reference
line, with a minimum turning radius on its edges.

\subsection{Trajectory methods: when to be where}\label{sec:traj}

The four trajectory methods are shown in \cref{fig:methods}e--h. 

The \textbf{Frenet lattice planner}
(lattice)~\cite{werling2010lattice} samples polynomials in path coordinates from
the measured state to several lateral offsets, durations and end speeds, and adds stopping and following
candidates. It returns the cheapest collision-free candidate; the cost covers comfort, speed and distance to the
path, plus penalties for proximity to agents and for leaving the drivable area.

The \textbf{optimization planner} (optim) optimizes the controls of a kinematic bicycle model over a 3\,s horizon,
starting at the measured speed:
\begin{equation}
  \min_{\vect x,\vect u}\ \textstyle\sum_{k} J_k\quad \text{s.t.}\ \ \vect x_{k+1}=f(\vect x_k,\vect u_k).
  \label{eq:nlp}
\end{equation}
The running cost $J_k$ has the same terms as the lattice's. Because the problem is non-convex, we run the gradient-based
solver from several initial guesses (the path, copies shifted to the left and right, and a braking profile) and
return the collision-free solution with the lowest cost.

The \textbf{$s$--$t$ search}~\cite{fan2018apollo} keeps the path fixed and chooses only how far along
it the ego travels over time. It works on a distance--time graph, a grid of the distance $s$ along the path against
the time $t$. For every agent, it blocks the cells in which the ego would come too close to that agent's forecast,
and then searches for the cheapest speed profile through the remaining cells. Depending on where the blocked cells
lie, this profile follows the agent, yields to it, passes before it, or stops.

The \textbf{dynamic window approach} (DWA)~\cite{fox1997dwa} considers the speeds and yaw rates that the car can
reach within the next 0.5\,s and predicts the arc it would drive with each. It discards every arc on which the car
could no longer stop before an obstacle and picks the one that best balances heading
towards the path, distance to obstacles and speed.

%% file: sections/04_protocol.tex
\section{Benchmarks and Tuning}\label{sec:protocol}

\textbf{nuScenes.} On the validation scenes of nuScenes~\cite{caesar2020nuscenes}, the planners run open loop on
UniAD's perception. We report two common metrics. The metric of UniAD takes the L2 error to the human trajectory at 1, 2 and 3\,s and tests collisions with a
coarse, axis-aligned box. The metric of Li et~al.~\cite{li2024egostatus} averages the error up to each horizon, which gives
smaller values, and tests the true ego box at every step. Values from the two are not comparable.

\textbf{nuPlan.} On nuPlan~\cite{caesar2021nuplan}, the planners run closed loop on ground-truth objects with
constant-velocity forecasts and the HD map, whose route centerline replaces the reference line (path source
\emph{route}). We report the closed-loop score on Val14~\cite{dauner2023pdm} and Test14-hard~\cite{ch01:cheng2024plantf} with
non-reactive (NR) and reactive (R) agents.

\textbf{HUGSIM.} HUGSIM~\cite{zhou2024hugsim} renders camera images from Gaussian-splatting reconstructions of real
drives (nuScenes, Waymo, PandaSet, KITTI-360) and inserts cars that stand, drive in the same direction, come towards the ego, cross or steer at
the ego. Its 436 scenarios fall into four tiers, from easy (no inserted car) to extreme (mostly cars that steer at the
ego). A run starts at 1\,m/s and ends at the first collision, when the ego leaves or completes the route, or after about 100\,s. The
HD-Score (0--100) is the route completion times the mean per-step score for collisions, drivable area, time to
collision and comfort. We average it with equal weight for each source dataset and tier, and report gains over UniAD's planner as
per-scenario differences, averaged the same way, $\pm$ one standard error. We use the
official UniAD v2.0 checkpoint. In our setup, UniAD's planner, i.e., its final plan after its own collision optimizer, scores 32.0, in line with an
independent evaluation~\cite{ch08:wajepa2026}. For the textbook planners, UniAD still runs completely on every
frame, and only its final plan is replaced. 
% We use UniAD because it decodes the outputs that textbook planners need, and
% HUGSIM because its real-world renders let the nuScenes-trained checkpoint run without retraining.

\textbf{Parameter tuning.} We tune the cost weights and clearance margins of each planner by random search: on
nuScenes, on 120 training scenes against the sum of mean L2 and mean collision rate; on nuPlan, on 84 training
scenarios against the official non-reactive score. HUGSIM received no tuning: the nuScenes configurations were copied
unchanged.

%% file: sections/05_results.tex
\section{Results on nuScenes and nuPlan}\label{sec:results}

\input{sections/tab_nusc}
\input{sections/tab_nuplan}
\input{sections/tab_all16}

\textbf{nuScenes: low L2, few collisions.} 
On UniAD’s metric, a lattice on the reference line reaches an L2 error of 0.735 m against 0.947 m for UniAD’s planner (\cref{tab:nusc}). This matches ReAL-AD [40] (0.74 m), a recent learned model that builds on UniAD. 
Part of this is expected: the textbook planners start from the measured
speed and acceleration, which open-loop L2 rewards~\cite{ch01:zhai2023admlp,li2024egostatus}, and learned planners
that receive the ego status reach an even lower L2 (0.35\,m against 0.43\,m on the metric of Li et al.). The
collision rate depends on the metric. On the stricter test of Li et al., the same planner collides  less often than
UniAD's planner (0.14\,\% against 0.53\,\%). 
On UniAD's coarse
test, it collides more often (0.75\,\% against 0.28\,\%); most of these collisions come from two scenes with a
standing ego, where the coarse test misplaces the ego box. Across the sixteen planners, the trajectory method matters more than the path source: the lattice has the lowest
errors, and DWA, which rolls out constant arcs, the highest.

\textbf{nuPlan: close to PDM-Closed.} With ground-truth objects and the map route, a lattice on the
route scores 90.0 on Val14, 2.8 points below PDM-Closed (\cref{tab:nuplan}). On Test14-hard with reactive agents, it
scores 79.3, second only to Diffusion Planner with refinement among the planners in \cref{tab:nuplan}. Test14-hard
was built from scenarios that are hard for PDM-Closed~\cite{ch01:cheng2024plantf}.
Among the path sources, the route is best for every trajectory method except the $s$--$t$ search
(\cref{tab:all16}): searched paths cut corners through the oncoming lane, which nuPlan penalizes. Unlike PDM-Closed, our planners select trajectories
with their own costs, not with a copy of the benchmark metric. No planner is best everywhere: hybrid A* + optim, the
best on HUGSIM, reaches 0.871\,m on nuScenes and 81.6 on Val14.

\begin{figure*}[t]
\centering
\includegraphics[width=\textwidth]{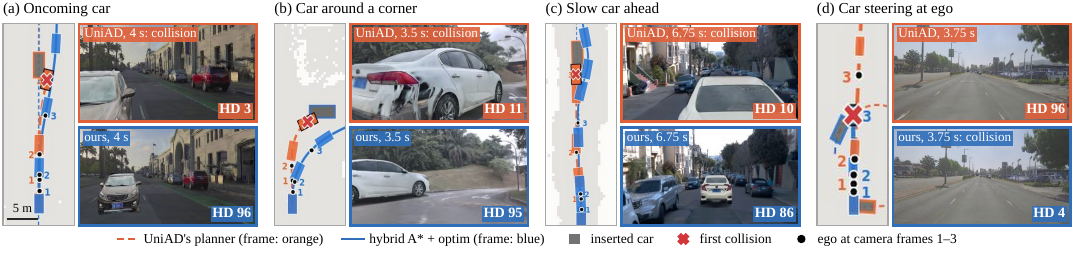}
\caption{Four HUGSIM scenarios on the same UniAD perception: bird's-eye view of both runs (UniAD's planner orange,
dashed; hybrid A* + optim blue) and the front camera at the decisive moment, with the HD-Score of each run. (a)~A car
approaches in the ego lane; UniAD's planner, faster, hits it head-on, while the textbook planner arrives later and
further right. (b)~A car stands just past a right turn; UniAD's planner turns late and hits it. (c)~A car drives ahead
at 2\,m/s; UniAD's planner rear-ends it, the textbook planner follows. (d)~A car steers at the ego; UniAD's planner
accelerates away, and the slower textbook planner is hit.}
\label{fig:examples}
\end{figure*}

\section{Results on HUGSIM with UniAD's Perception}\label{sec:hugsim}

\input{sections/tab_hugsim}

\textbf{Does a textbook planner drive better?} Sometimes. On the same perception, nine of the sixteen planners score
above UniAD's planner and seven below (\cref{tab:hugsim}). Six of the gains and two of the losses exceed two standard
errors. The two largest gains are both optim on a searched path: hybrid A* + optim gains $9.7\pm1.4$ points and
RRT* + optim $9.0\pm1.4$, so the result is not one lucky configuration. The combination matters: optim is clearly ahead only on the two searched paths with bounded curvature (hybrid A*,
RRT*). The $s$--$t$ search is weakest (29.4 on average): with its tuned weights, it barely accelerates from the
start speed and completes little of the route. All sixteen planners collide
less often per driven step than UniAD's planner, but only the two best also complete more of the route; the others
drive more slowly and get less far.

\input{sections/tab_hugsim_pub}

\textbf{How does it compare with published models?} \Cref{tab:hugsimpub} lists published HUGSIM results. On the same
436 scenarios, as evaluated in~\cite{ch08:wajepa2026}, the best textbook planner (41.8) is ahead of VAD, LTF, UniAD
(32.1) and DrivoR (35.3), and behind only the world-action model WA-JEPA (44.9). Results
from other simulator versions differ in scenario set and scoring: UniAD alone scores between 25.7 and 34.5 across
them~\cite{drivezero2026,terratransferpp2026}. They are therefore only indicative: the best textbook planner is on
par with DriveZero (42.6), ahead of GigaPixel and DrivoR-Scale, and behind DriveZero-Scale, trained with additional
simulated data, and TerraTransfer++, which adapts to HUGSIM's renders.

\input{sections/tab_actor}

\textbf{Where does the gain come from?} We now compare the best planner, hybrid A* + optim, with UniAD's planner (\cref{tab:actor}). The gain is largest with standing and oncoming cars, the situations that
the two stages of a textbook planner are built for. Unless UniAD detects it straight ahead in the lane, a standing car is an obstacle on the path source's grid, so the
path goes around it (\cref{fig:teaser}b, \cref{fig:examples}b): UniAD's planner hits the standing car in 31 of 80
scenarios, the textbook planner in 13. A moving car is checked against its forecast at every time step, so the
trajectory makes room, waits or follows (\cref{fig:examples}a,c). The tiers show the same pattern
(\cref{tab:hugsimpub}): the gain is largest in the medium and hard tiers ($+14.8$ and $+13.1$), which contain the
standing, slow and oncoming cars, and small in the extreme tier ($+3.4$), where most cars steer at the ego. The gain
also holds on all four source datasets: $+11.9$, $+12.7$ and $+10.8$ on nuScenes, Waymo and PandaSet scenes, and
$+6.5$ on KITTI-360, where both planners score low. 
% How the runs end confirms it: 55\,\% of the textbook planner's
% runs end in a collision with a car or the static scene against 66\,\% for UniAD's planner, and it completes the
% route in 158 of 435 runs (UniAD's planner: 120).

\textbf{Perception or planner?} UniAD's planner hits a car in 182 runs. In 31 of them, the textbook planner completes
the route on the same perception: these crashes are due to the planner, not the perception. In another 131, both
planners hit a car, and we cannot tell which part is to blame.

\textbf{Is the gain just caution?} No. The textbook planner starts more slowly: from the common start at 1\,m/s, it
reaches 2.2\,m/s after 1\,s against 3.5\,m/s for UniAD's planner, which crashes within the first 4\,s in 125 runs
against 70. But a slow start alone does not explain the gain: it is smallest in the 133 scenarios where the textbook
planner is more than 1.5\,m/s slower than UniAD's planner over the first 4\,s ($+4.0$) and largest in the 110
where the two speeds differ by less than 0.5\,m/s ($+14.6$).

\textbf{Where does textbook planning lose?} Against cars that steer at the ego, there is no gain (\cref{tab:actor}).
UniAD's forecasts do not anticipate that such a car turns towards the ego, so the outcome is a matter of timing: in
the largest losses, UniAD's faster planner has already passed when the car arrives, and the slower textbook planner
is hit
(\cref{fig:examples}d). The textbook planner also leaves the route more often in turns. Without an HD map, it depends on UniAD's predicted road masks, which often miss the branch that the
route takes.

%% file: sections/tab_nusc.tex
\begin{table}[t]
\centering
\caption{nuScenes, open loop: L2  and collision rate. Bold: best.}
\label{tab:nusc}
\footnotesize
\setlength{\tabcolsep}{4pt}
\begin{tabular}{@{}lcccc@{}}
\toprule
 & \multicolumn{2}{c}{UniAD metric} & \multicolumn{2}{c}{metric of Li et al.} \\
\cmidrule(lr){2-3}\cmidrule(l){4-5}
planner & L2 & Col & L2 & Col \\
\midrule
UniAD v1.0~\cite{hu2023uniad,li2024egostatus}        & 1.03 & 0.31 & 0.66 & 0.62 \\
ReAL-AD~\cite{ch01:realad}                        & \textbf{0.74} & \textbf{0.17} & -- & -- \\
Ego-MLP~\cite{li2024egostatus}                        & -- & -- & \textbf{0.35} & 0.37 \\
\midrule
UniAD's planner (v2.0, our run)  & 0.947 & 0.277 & 0.648 & 0.534 \\
reference line + lattice (ours) & \textbf{0.735} & 0.753 & 0.430 & \textbf{0.137} \\
hybrid A* + optim (ours)         & 0.871 & 0.792 & 0.498 & 0.241 \\
\bottomrule
\multicolumn{5}{@{}l@{}}{}
\end{tabular}
\end{table}

%% file: sections/tab_nuplan.tex
\begin{table}[t]
\centering
\caption{nuPlan closed-loop score with ground-truth perception, with non-reactive (NR) and reactive (R) agents. Bold: best.}
\label{tab:nuplan}
\footnotesize
\setlength{\tabcolsep}{3.5pt}
\begin{tabular}{@{}l cc cc@{}}
\toprule
 & \multicolumn{2}{c}{Val14} & \multicolumn{2}{c}{Test14-hard} \\
\cmidrule(lr){2-3}\cmidrule(l){4-5}
planner & NR & R & NR & R \\
\midrule
PDM-Closed~\cite{dauner2023pdm} (rule-based)  & 92.8 & 92.1 & 65.1 & 75.2 \\
PLUTO~\cite{ch01:cheng2024pluto} (learned + rules) & 92.9 & 89.7$^a$ & \textbf{80.1} & 76.9 \\
Diff.\ Planner + refine~\cite{diffusionplanner2025} (learned + rules) & \textbf{94.3} & \textbf{92.9} & 78.9 & \textbf{82.0} \\
Diff.\ Planner~\cite{diffusionplanner2025} (learned) & 89.9 & 82.8 & 76.0 & 69.2 \\
CaRL~\cite{jaeger2025carl} (learned, RL)      & 91.3 & 90.6 & -- & -- \\
\midrule
route + lattice (ours)                        & 90.0 & 90.4 & 75.3 & 79.3 \\
hybrid A* + optim (ours)                      & 81.6 & 84.0 & 65.6 & 72.9 \\
\bottomrule
\multicolumn{5}{@{}p{0.95\columnwidth}@{}}{\scriptsize Published values from~\cite{diffusionplanner2025}; CaRL
and the value marked $^a$ from~\cite{jaeger2025carl}.}
\end{tabular}
\end{table}

%% file: sections/tab_all16.tex
\begin{table}[t]
\centering
\caption{All sixteen planners on nuScenes (metrics of UniAD and of Li et al.: L2, collision rate) and
nuPlan (Val14 NR, Test14-hard R). Bold: best.}
\label{tab:all16}
\footnotesize
\setlength{\tabcolsep}{3.2pt}
\begin{tabular}{@{}llccccc@{}}
\toprule
 & & UniAD & \multicolumn{2}{c}{Li et al.} & \multicolumn{2}{c}{nuPlan} \\
\cmidrule(lr){3-3}\cmidrule(lr){4-5}\cmidrule(l){6-7}
path source & trajectory & L2 & L2 & Col & Val14 & T14-hard \\
\midrule
\multicolumn{2}{@{}l}{UniAD's planner} & 0.947 & 0.648 & 0.53 & -- & -- \\
\midrule
\multirow{4}{*}{reference line} & lattice & \textbf{0.735} & \textbf{0.430} & \textbf{0.14} & \textbf{90.0} & \textbf{79.3} \\
 & optim & 0.893 & 0.507 & 0.27 & 86.6 & 74.8 \\
 & $s$--$t$ & 0.949 & 0.540 & 0.26 & 82.4 & 70.4 \\
 & DWA & 1.273 & 0.782 & 0.53 & 87.7 & 76.8 \\
\addlinespace[2pt]
\multirow{4}{*}{A*} & lattice & 0.832 & 0.507 & 0.24 & 76.0 & 62.6 \\
 & optim & 0.912 & 0.533 & 0.25 & 81.0 & 68.2 \\
 & $s$--$t$ & 1.140 & 0.694 & 0.24 & 75.0 & 62.8 \\
 & DWA & 1.341 & 0.817 & 0.53 & 87.2 & 76.6 \\
\addlinespace[2pt]
\multirow{4}{*}{hybrid A*} & lattice & 0.783 & 0.481 & 0.18 & 78.5 & 66.5 \\
 & optim & 0.871 & 0.498 & 0.24 & 81.6 & 72.9 \\
 & $s$--$t$ & 1.156 & 0.697 & 0.27 & 74.6 & 60.7 \\
 & DWA & 1.528 & 0.936 & 0.52 & 83.3 & 74.7 \\
\addlinespace[2pt]
\multirow{4}{*}{RRT*} & lattice & 0.790 & 0.467 & 0.15 & 82.8 & 75.9 \\
 & optim & 0.858 & 0.490 & 0.23 & 81.1 & 74.8 \\
 & $s$--$t$ & 1.226 & 0.705 & 0.23 & 83.6 & 70.6 \\
 & DWA & 1.350 & 0.831 & 0.46 & 83.4 & 74.8 \\
\bottomrule
\end{tabular}
\end{table}

%% file: sections/tab_hugsim.tex
\begin{table}[t]
\centering
\caption{HUGSIM HD-Score of the sixteen textbook planners on UniAD's perception, with the gain over UniAD's planner
(32.0) in brackets. The standard error of each gain is 1.4--1.6; bold: gain above two standard
errors, underlined: loss beyond two.}
\label{tab:hugsim}
\footnotesize
\renewcommand{\arraystretch}{1.25}
\setlength{\tabcolsep}{3pt}
\begin{tabular}{@{}lcccc@{}}
\toprule
path source & lattice & optim & $s$--$t$ search & DWA \\
\midrule
reference line & 34.9 (+2.8) & \textbf{35.6 (+3.6)} & 29.5 ($-$2.5) & 31.5 ($-$0.5) \\
A* & \textbf{35.9 (+3.9)} & 33.8 (+1.8) & \underline{26.2 ($-$5.8)} & \textbf{35.7 (+3.7)} \\
hybrid A* & \textbf{35.1 (+3.1)} & \textbf{41.8 (+9.7)} & 30.7 ($-$1.3) & 33.9 (+1.9) \\
RRT* & 30.7 ($-$1.3) & \textbf{41.0 (+9.0)} & 31.2 ($-$0.8) & \underline{28.0 ($-$4.0)} \\
\bottomrule
\end{tabular}
\end{table}

%% file: sections/tab_hugsim_pub.tex
\begin{table}[t]
\centering
\caption{HUGSIM HD-Score by tier, compared with published methods. Mean: unweighted mean of the four tiers,
recomputed by us.}
\label{tab:hugsimpub}
\footnotesize
\setlength{\tabcolsep}{3pt}
\begin{tabular}{@{}lccccc@{}}
\toprule
method & easy & medium & hard & extreme & mean \\
\midrule
\multicolumn{6}{@{}l}{\emph{same 436 scenarios and simulator}} \\
\textbf{hybrid A* + optim (ours)} & 71.0 & 48.3 & 34.8 & 13.0 & 41.8 \\
UniAD's planner (our run) & 63.2 & 33.5 & 21.7 & 9.6 & 32.0 \\
WA-JEPA~\cite{ch08:wajepa2026} & 79.8 & 55.6 & 30.6 & 13.6 & 44.9 \\
DrivoR~\cite{kirby2026drivor,ch08:wajepa2026} & 78.0 & 29.1 & 20.0 & 14.1 & 35.3 \\
UniAD~\cite{hu2023uniad,ch08:wajepa2026} & 64.0 & 37.2 & 21.0 & 6.3 & 32.1 \\
LTF~\cite{ch01:dauner2024navsim,ch08:wajepa2026} & 66.1 & 15.5 & 12.0 & 11.7 & 26.3 \\
VAD~\cite{ch01:jiang2023vad,ch08:wajepa2026} & 42.0 & 8.5 & 7.7 & 6.3 & 16.1 \\
\midrule
\multicolumn{6}{@{}l}{\emph{other simulator versions or scenario sets (indicative)}} \\
TerraTransfer++$^a$~\cite{terratransferpp2026} & 82.1 & 71.2 & 49.3 & 16.1 & 54.7 \\
DriveZero-Scale~\cite{drivezero2026} & 81.9 & 60.5 & 24.7 & 27.6 & 48.7 \\
DriveZero~\cite{drivezero2026} & 78.5 & 47.2 & 16.0 & 28.7 & 42.6 \\
GigaPixel~\cite{drivezero2026} & 67.4 & 51.9 & 19.1 & 21.6 & 40.0 \\
DrivoR-Scale~\cite{drivezero2026} & 64.8 & 50.3 & 16.8 & 25.8 & 39.4 \\
\bottomrule
\multicolumn{6}{@{}p{0.95\columnwidth}@{}}{\scriptsize $^a$Uses the end point of the logged route as its goal.}
\end{tabular}
\end{table}

%% file: sections/tab_actor.tex
\begin{table}[t]
\centering
\caption{HUGSIM by type of inserted car: plain mean of the HD-Score over the scenarios of each type (hence slightly
different from Tables~\ref{tab:hugsim} and~\ref{tab:hugsimpub}), and gain $\pm$ one standard error (bold: above two
standard errors).}
\label{tab:actor}
\footnotesize
\setlength{\tabcolsep}{6pt}
\begin{tabular}{@{}lccc@{}}
\toprule
inserted car & UniAD's planner & hybrid A* + optim & gain \\
\midrule
standing & 26.6 & 45.4 & \textbf{+18.8$\pm$4.2} \\
oncoming & 12.5 & 30.3 & \textbf{+17.8$\pm$4.1} \\
crossing & 22.0 & 37.1 & +15.1$\pm$8.0 \\
same direction & 44.0 & 54.9 & \textbf{+10.8$\pm$3.4} \\
several types & 21.1 & 30.9 & +9.8$\pm$5.4 \\
none & 59.3 & 67.4 & \textbf{+8.1$\pm$3.0} \\
steers at ego & 13.9 & 12.9 & $-$1.0$\pm$2.8 \\
\bottomrule
\end{tabular}
\end{table}

%% file: sections/07_discussion.tex
\section{Discussion}\label{sec:discussion}

\textbf{Implications.} Textbook planners are not a thing of the past: the best are competitive on nuScenes and nuPlan
and, on the perception of an end-to-end model, drive better than its learned planner. For end-to-end models with
interpretable outputs, a textbook planner on the model's own perception is a cheap and informative baseline: because both planners read
the same perception, it separates errors of the planner from errors of the perception. The
failure modes are partly complementary, so learned proposals could be checked with an explicit, time-indexed
collision test, as rule-based selection does on nuPlan~\cite{dauner2023pdm,ch01:cheng2024pluto}.

\textbf{Limitations.} All results on learned perception use one end-to-end model; models with stronger learned
planners may behave differently. We did not run the planners on the simulator's true boxes, so the runs in which
both planners hit a car cannot be divided between perception and planning. HUGSIM's scenarios are variants of a fixed set of
base scenes and thus not independent, as our standard errors assume. HUGSIM inserts only cars, and its score mostly judges the planned
waypoints, not the driven motion. It also ignores the driving direction, so using the oncoming lane costs nothing. 

\textbf{Conclusion.} Old-school planners sometimes drive better than learned ones. The best textbook planners are
competitive on nuScenes and nuPlan, and on UniAD's unchanged perception the best of them clearly beats UniAD's
planner in closed loop. They still fail in interactive scenarios: against cars that steer at the ego, where the
forecasts they rely on are wrong, they gain nothing. Combining rule-based and learned planning is therefore a
promising direction, as the strongest planners on nuPlan already
show~\cite{ch01:cheng2024pluto,diffusionplanner2025}. Repeating the substitution with other end-to-end models, and
with the simulator's true boxes to separate perception from planning errors, is future work.